\documentclass[11pt]{article}
\usepackage{fullpage}
\usepackage{amssymb,calc,enumerate,booktabs}
\usepackage{amsmath,mathrsfs,yfonts}
\usepackage{amsthm}
\usepackage{etoolbox,url,subfigure}
\usepackage[ruled,linesnumbered]{algorithm2e}
\usepackage{units}
\usepackage{tikz,adjustbox}
\usepackage{pgfplots}
\usepackage[affil-sl]{authblk}
\usepackage{multirow}
\usepackage{threeparttable}
\usepackage{rotating}
\newcommand{\cl}[1]{\mathscr{#1}}

\makeatletter
\def\@maketitle{%
  \newpage
  \null
  \vskip 2em%
  \begin{center}%
  \let \footnote \thanks
    {\Large\bfseries \@title \par}%
    \vskip 1.5em%
    {\normalsize
      \lineskip .5em%
      \begin{tabular}[t]{c}%
        \@author
      \end{tabular}\par}%
    \vskip 1em%
    {\normalsize \@date}%
  \end{center}%
  \par
  \vskip 1.5em}
\makeatother

 \newtheoremstyle{component}{}{}{}{}{\bfseries\itshape}{:}{.5em}{\thmnote{#3}#1}
    \theoremstyle{component}
    
\makeatletter
\patchcmd{\@algocf@start}{%
  \begin{lrbox}{\algocf@algobox}%
}{%
  \rule{0.1\textwidth}{\z@}%
  \begin{lrbox}{\algocf@algobox}%
  \begin{minipage}{0.8\textwidth}%
}{}{}
\patchcmd{\@algocf@finish}{%
  \end{lrbox}%
}{%
  \end{minipage}%
  \end{lrbox}%
}{}{}
\newenvironment{mathprog}[1][]%
{\par\bigskip\noindent\ignorespaces\begin{tabular*}{\textwidth}{@{} p{0.35\textwidth} @{\hspace{0.5em}} l} #1}%
{\end{tabular*}\\\smallskip\par\noindent\ignorespacesafterend}

\newcommand{\progline}[2][]{\hfill #1 & \begin{math}\displaystyle #2 \end{math}\\}

\makeatother
\title{\LARGE Integrating tabu search and VLSN search to develop enhanced algorithms: A case study using bipartite boolean quadratic programs\footnote{This research work was supported by an NSERC  Discovery grant and an NSERC discovery accelerator grant awarded to Abraham P Punnen. }}
\author {Fred Glover\thanks{glover@opttek.com}}
\affil{OptTek Systems, Boulder, Colorado, USA}
\author {Tao Ye\thanks{yeetao@gmail.com}}
\author {Abraham P. Punnen\thanks{apunnen@sfu.ca}}
\affil{Department of Mathematics, Simon Fraser University,  Surrey, British Columbia, V3T 0A3, Canada}
\author {Gary Kochenberger\thanks{Gary.Kochenberger@cudenver.edu}}
\affil{School of Business, University of Colorado at Denver, Denver, Colorado , USA}

\date{}

\begin{document}

\maketitle

\begin{abstract}
The bipartite boolean quadratic programming problem (BBQP) is a generalization of the well studied boolean quadratic programming problem. The model has a variety of real life applications; however, empirical studies of the model are not available in the literature, except in a few isolated instances. In this paper, we develop efficient heuristic algorithms based on tabu search, very large scale neighborhood (VLSN) search,  and a hybrid algorithm that integrates the two. The computational study establishes that effective integration of simple tabu search with VLSN search results in superior outcomes, and suggests the value of such an integration in other settings. Complexity analysis and implementation details are provided along with conclusions drawn from experimental analysis. In addition, we obtain solutions better than the best previously known for almost all medium and large size benchmark instances.\\

\end{abstract}

\section{Introduction}

Local search algorithms and their metaheuristic elaborations such as tabu search have become the methods of choice for solving many complex applied optimization problems. Traditional local search algorithms use exhaustive search  over small neighborhoods while many recent local search algorithms use neighborhoods of exponential size  that often can be searched for an improving solution in polynomial time. To distinguish between these variations, the former is called {\it simple neighborhood search} (SN search) and the latter is called {\it very large-scale neighborhood search} (VLSN search)\cite{vlsn}. SN search based local search algorithms are generally faster in exploring neighborhoods but take a large number of iterations to reach a locally optimal solution. Many VLSN search algorithms on the other hand  take longer to search a neighborhood for an improving solution but often reach a locally optimal solution quickly within a relatively small number of iterations. In this paper we consider an integration of SN search and VLSN search within a tabu search framework to develop enhanced algorithms  for an important combinatorial optimization problem called the {\it   bipartite boolean quadratic programming problem} (BBQP).

Let $Q = (q_{ij})$ be an $m\times n$ matrix, $c=(c_1,c_2, \ldots ,c_m)$ be a row vector in $R^m$ and $d=(d_1,d_2,\ldots ,d_n)$ be a row vector in $R^n$\@. Then, the problem (BBQP) can be stated mathematically as
\begin{mathprog}
\progline[BBQP:  Maximize]{f(x,y) = x^TQy + cx+dy}
\progline[subject to]{x \in \{0,1\}^m, y\in \{0,1\}^n.}
\end{mathprog}

 An instance of BBQP is completely defined by the matrix $Q$ and vectors $c$ and $d$ and hence it is represented by $\cl{P}(Q,c,d)$. BBQP has been studied by many researchers in various applications including clustering and bioinformatics~\cite{tanay}, matrix factorization~\cite{gills,shen}, data mining~\cite{lu,shen}, solving basic graph theoretic optimization problems~\cite{amb,tan,p1}, and computing approximations to the cut-norm of a matrix~\cite{alon}. The problem can also be viewed as a generalization of the  well-studied {\it boolean quadratic programming problem} (BQP)~\cite{fred1998,fred2010,lu2010,wang2012}
\begin{mathprog}
\progline[BQP:  Maximize]{ f(x) = x^TQ^{\prime}x + c^{\prime}x}
\progline[subject to]{ x \in \{0,1\}^n,}
\end{mathprog}
where $Q^{\prime}$ is an $n\times n$ matrix and $c{^{\prime}}$ is a row vector in $R^n$. As pointed out in~\cite{p1} and \cite{abraham2012b},
 by choosing
 \begin{equation}\label{equa}Q=Q^{\prime}+2MI,\; c=\frac{1}{2}c^{\prime}-Me \mbox{ and } d=\frac{1}{2}c^{\prime}-Me,\end{equation}
 where $I$ is an $n\times n$ identity matrix, $e\in R^n$ is an all one vector and $M$ is a very large number, BQP can be formulated as a BBQP.  Thus, systematic experimental study of algorithms for  BBQP is also relevant for all the applications studied in the context of BQP.

 Despite its unifying role and various practical applications, BBQP has not been investigated thoroughly from the experimental analysis point of view. The only systematic study that we are aware of is by Karapetyan and Punnen~\cite{kp} who generated a class of test instances and provided experimental results with various heuristic algorithms. Some limited experimental study using a specific algorithm, called the alternating algorithm~\cite{kp,lu} is also available in the context of specific applications.

 This paper focuses on developing efficient heuristic algorithms for solving BBQP. We present two neighborhood structures, a classic one-flip neighborhood and a new flip-float neighborhood, and based on them propose a one-flip move based tabu search algorithm, a flip-float move based coordinate method and a hybrid algorithm that combines the two.  While the specific optimization problem addressed in this paper is BBQP, our approach for integrating tabu search (TS) and VLSN search is applicable to other settings to obtain hybrid algorithms that inherit  individual properties of these algorithmic paradigms. Computational experiments on a set of 85 benchmark instances ~\cite{kp} disclose that the hybrid TS/VLSN method shows better performance in terms of both  solution quality and robustness than either of its component methods in isolation. The hybrid method is able to improve almost all the previous best-known solutions on the medium and large size instances.

 We also compared our algorithms with a ready-made heuristic that solves an integer programming formulation of BBQP using CPLEX~\cite{ibm} with appropriate parameter settings guiding the solver to emphasize on producing a heuristic solution within a prescribed time limit. This approach produced solutions comparable to that of our algorithms (but at the cost of additional time) for small size instances. For medium and large size instances, this approach proved to be impractical, reinforcing the need for special purpose algorithms to solve BBQP.

 We also provide landscape analysis of the benchmark instances to identify the inherent difficulty of these instances for local search algorithms. Such a study is not available in literature so far and this provides additional insights into the structure of the benchmark instances.

 The rest of the paper is organized as follows. Section 2 presents two neighborhood structures and describes in detail three heuristic algorithms.  Section 3 reports and discusses computational statistics of the proposed algorithms on the standard benchmark instances. We also discuss landscape analysis of the benchmark instances in this section.  Finally, concluding remarks are provided in Section 4.

\section{Neighborhoods and heuristics algorithms}
This section proposes three heuristic algorithms for solving BBQP. The first algorithm adopts  a classic one-flip neighborhood structure and a  tabu search strategy. The second algorithm employs a new  flip-float neighborhood structure and a coordinate ascent strategy. The third  integrates the first and second algorithms to produce a hybrid method.

\subsection{One-flip neighborhood}
The classic one-flip move performed on a binary vector consists of changing the value of a component of the vector to its complementary value (i.e., flipping a component of the vector). Thus, by flipping the $i$-th component of vector $x=(x_1, x_2, \dots, x_m)$,  we get a new vector $x'=(x_1, \dots,x_{i-1}, 1-x_i, x_{i+1}, \dots, x_m)$. This flip process can be denoted as:
\begin{equation}
x' \leftarrow x \oplus Flip(i).
\end{equation}
For a solution $(x, y)$ of BBQP, we can perform one-flip moves on both $x$ and $y$ and hence the number of all possible one-flip moves is $m+n$.

In local search algorithms, we usually need to rapidly determine the effect of a one-flip move on the objective function. To achieve this, we adapt a fast incremental evaluation technique, widely used for the BQP problem \cite{fred1998,fred2010,fred2010b,lu2010,wang2011,wang2012}, to the BBQP problem. Specifically, we use two arrays to store the contribution of each possible move, and employs a streamlined calculation for updating the arrays after each move.

Let $\Delta x_i$ $(i=1, 2, \dots, m)$ denote the change in the objective function value caused by flipping the $i$-th component of $x$ and let $x' = x \oplus Flip(i)$, then
\begin{equation}
\label{eq:deltaxi}
\Delta {x_i}  = f(x', y) - f(x,y) =  (x'_i - x_i)(c_i + \sum_{j=1}^n y_jq_{ij}).
\end{equation}
Similarly, let $\Delta y_j$ $(j=1, 2, \dots, n)$ denote the change in the objective function value caused by flipping the $j$-th component of $y$ and let $y' = y \oplus Flip(j)$. Then
\begin{equation}
\label{eq:deltayj}
\Delta {y_j}  = f(x, y') - f(x,y) = (y'_j - y_j)(d_j + \sum_{i=1}^m x_iq_{ij}).
\end{equation}

In our implementation, we use two arrays to store all the $\Delta{x_i}$ and $\Delta{y_j}$ values. These arrays can be initialized using equations (\ref{eq:deltaxi}) and (\ref{eq:deltayj}) with time complexity $O(mn)$. After a move is performed, the $\Delta x_i$ and $\Delta y_j$ arrays are updated efficiently. Note that we just need to update the elements affected by the move and the new values can be determined incrementally. A detailed description of the algorithm for updating the $\Delta{x_i}$  and $\Delta{y_j}$ values with time complexity $O(m+n)$ is given in Algorithm \ref{alg:update}.
\begin{algorithm}
\caption{The $\Delta x_i$  and $\Delta y_j$ arrays updating algorithm}
\label{alg:update}
\If{the move is $x_i: 1 \rightarrow 0$}
{
   $\Delta x_i \leftarrow - \Delta x_i$ \;
   \For{$j \leftarrow 1$ to $n$}
   {
      \eIf{$y_j=1$}
      {
         $\Delta y_j \leftarrow \Delta y_j + q_{ij}$ \;
      }
      {
          $\Delta y_j \leftarrow \Delta y_j - q_{ij}$ \;
      }

   }
}

\If{the move is $x_i: 0 \rightarrow 1$}
{
   $\Delta x_i \leftarrow - \Delta x_i$ \;
   \For{$j \leftarrow 1$ to $n$}
   {
      \eIf{$y_j=1$}
      {
         $\Delta y_j \leftarrow \Delta y_j - q_{ij}$ \;
      }
      {
          $\Delta y_j \leftarrow \Delta y_j + q_{ij}$ \;
      }

   }
}
\If{the move is $y_j: 1 \rightarrow 0$}
{
   $\Delta y_j \leftarrow - \Delta y_j$ \;
   \For{$i \leftarrow 1$ to $m$}
   {
      \eIf{$x_i=1$}
      {
         $\Delta x_i \leftarrow \Delta x_i + q_{ij}$ \;
      }
      {
          $\Delta x_i \leftarrow \Delta x_i - q_{ij}$ \;
      }
   }
}

\If{the move is $y_j: 0 \rightarrow 1$}
{
   $\Delta y_j \leftarrow - \Delta y_j$ \;
   \For{$i \leftarrow 1$ to $m$}
   {
      \eIf{$x_i=1$}
      {
         $\Delta x_i \leftarrow \Delta x_i - q_{ij}$ \;
      }
      {
          $\Delta x_i \leftarrow \Delta x_i + q_{ij}$ \;
      }

   }
}

\end{algorithm}

\subsection{One-flip move based tabu search algorithm}
Based on the one-flip neighborhood, we can formulate a simple tabu search algorithm as described in Algorithm \ref{alg:tabu}. When a component of $x$ or $y$ is flipped in the current step, it is declared tabu for the next $TabuTenure$ steps, i.e., one-flip moves (either from 0 to 1 or from 1 to 0) involving this component are forbidden for the next $TabuTenure$ iterations.  Using a simple default setting, the $TabuTenure$ for $x_i$ $(i=1,2 \dots, m)$ is set to $m/20 + rand(0,10)$, where $rand(0,10)$ denotes a random integer between 0 and 10. For $y_j$ $(j=1,2,\dots,n)$, the $TabuTenure$ is set to $n/20 + rand(0,10)$. We also used a simple \textit{aspiration criterion}, allowing a tabu move to be performed if it leads to a solution better than the best-found one. Our rudimentary tabu search algorithm  starts from a randomly generated initial solution. In each iteration, it executes the best admissible one-flip move and repeats until the incumbent (best-found) solution has not been improved in the last $TabuDepth$ steps.

In the resulting algorithm, depicted as Algorithm \ref{alg:tabu} below, the time complexity of steps 1 and 2 is respectively $O(m+n)$ and $O(mn)$. Step 4 is realized by scanning the $\Delta x_i$ and $\Delta y_j$ arrays and looking up the TabuList with time complexity $O(m+n)$. Steps 5 and 6 can be done in $O(1)$ time. The complexity of Step 7 is $O(m+n)$, as indicated in the previous section. Therefore, for the one-flip move based tabu search algorithm, the complexity of each search step is $O(m+n)$.

\begin{algorithm}
\caption{The one-flip move based simple tabu search algorithm}
\label{alg:tabu}
\KwIn{An initial solution $(x, y)$}
\KwOut{The best solution found so far}
Initialize TabuList \;
Initialize the $\Delta x_i$ and $\Delta y_j$ arrays using Eq.\ref{eq:deltaxi} and Eq.\ref{eq:deltayj}\;

\Repeat{the best-found solution has not been improved in the last $TabuDepth$ iterations}
{
   Determine the best admissible move $mv$ by scanning the $\Delta x_i$ and $\Delta y_j$ arrays and looking up the TabuList \;

   Perform $mv$ \;

   Update TabuList \;
   Update $\Delta x_i$ and $\Delta y_j$ arrays using Algorithm \ref{alg:update} \;
}

\Return {The best solution found so far}
\end{algorithm}

\subsection{Flip-float neighborhood}
For a fixed $x=(x_1, x_2, \dots, x_m)$, we choose the best $y=y^*(x)$ which makes $f(x,y)$ maximal by the following equation \cite{kp}:
\begin{equation}
\label{eq:dyj}
y^*(x)_j = \left \{ \begin{array}{ll}
1 & \textrm{if} \  d_j + \sum_{i=1}^m x_iq_{ij} > 0; \\
0 & \textrm{otherwise.} \\
\end{array} \right.
\end{equation}

Similarly, we can choose the best $x=x^*(y)$ for a specific $y=(y_1, y_2, \dots, y_n)$ by the equation:
\begin{equation}
\label{eq:dyj}
x^*(y)_i = \left \{ \begin{array}{ll}
1 & \textrm{if} \  c_i + \sum_{j=1}^n y_jq_{ij} > 0; \\
0 & \textrm{otherwise.} \\
\end{array} \right.
\end{equation}

The {\it Flip-x-Float-y} move consists of flipping one component of $x$ and then choosing the best (floating) $y$ with respect to the flipped $x$. Similarly, the  {\it Flip-y-Float-x} move is defined as flipping one component of $y$ and then choosing the best (floating) $x$ with respect to the flipped $y$. The number of all possible {\it Flip-x-Float-y} and {\it Flip-y-Float-x} moves is, respectively, $m$ and $n$.

Let $F_{y^*}(x) = f(x, y^*(x))$, $F_{x^*}(y) = f(x^*(y), y)$. Then the change of objective function value caused by flipping the $i$-th component of $x$ and floating $y$ is given by:
\begin{equation}
\Delta \textrm{\it Flip-x-Float-y}(i) = F_{y^*}(x \oplus Flip(i)) - F_{y^*}(x).
\end{equation}
Similarly, the change in the objective function value caused by flipping the $j$th component of $y$ and floating $x$ is given by
\begin{equation}
\Delta \textrm{\it Flip-y-Float-x}(j) = F_{x^*}(y \oplus Flip(j)) - F_{x^*}(y).
\end{equation}

In the local search, we need to find a fast way to identify the value of $\Delta \textrm{\it Flip-x-Float-y}(i)$ and $\Delta\textrm{\it Flip-y-Float-x}(j)$. In the following, we only describe how to determine $\Delta\textrm{\it Flip-x-Float-y(i)}$. Computation of $\Delta\textrm{\it Flip-y-Float-x}(j)$ can be done in an analogous way by simply exchanging the symbols (1) $x$ and $y$, (2) $i$ and $j$, (3) $c_i$ and $d_j$, and (4) $m$ and $n$.

Let
\begin{equation}
\label{eq:sumxj}
Sum(x, j) = d_j + \sum_{i=1}^m{x_iq_{ij}}.
\end{equation} Then we have
\begin{equation}
\label{eq:sumyi}
F_{y^*}(x) = \sum_{i=1}^{m}c_ix_i + \sum_{j=1}^n{max(0, Sum(x, j))}
\end{equation}

Let $x'= x \oplus Flip(i)$ ($i=1, 2, \dots, m$). Then
\begin{eqnarray}
\label{eq:deltafyx}
F_{y^*}(x') - F_{y^*}(x) & = & \sum_{k=1}^{m}c_k {x'}_k + \sum_{j=1}^n{max(0, Sum(x', j))} \nonumber \\
                         &   &  - \sum_{k=1}^{m}c_i {x}_k - \sum_{j=1}^n{max(0, Sum(x, j))} \nonumber \\
                         & = &  c_i ({x'}_i - x_i) + \sum_{j=1}^{n}\large ( {max(0, Sum(x',j)) - max(0, Sum(x, j))} \large).
\end{eqnarray}

From Eq.\ref{eq:deltafyx}, we observe that computing the value of $\Delta \textrm{\it Flip-x-Float-y}(i)$ requires knowledge of the $Sum(x, j)$ values. Thus, in our calculations we use an array of size $n$ to store $Sum(x, j)$ $(j=1, 2, \dots, n)$. From Eq.\ref{eq:sumxj}, the $Sum(x,j)$ array can be initialized with $O(mn)$ time complexity. In each search step, we use the following algorithm (Algorithm \ref{alg:deltaF}) of complexity $O(n)$ to compute $\Delta \textrm{\it Flip-x-Float-y}(i)$. After performing move {\it Flip-x-Float-y}($i$), the $Sum(x,j)$ array can be updated by adding $({x'}_i - x_i)*q_{ij}$ to each element. It's clear that each updating requires $O(n)$ time complexity. Note that before applying {\it Flip-x-Float-y(i)}, we make sure that $y=y^*(x)$. Likewise, before applying {\it Flip-y-Float-x(j)}, we make sure $x=x^*(y)$.

\begin{algorithm}
\caption{Algorithm to compute  $\Delta \textrm{\it Flip-x-Float-y}(i)$}
\label{alg:deltaF}
\KwIn{$i$, $x$, $Sum(x, j)(j=1,2,\dots,n)$}
\KwOut{$\Delta \textrm{\it Flip-x-Float-y}(i)$}
$\Delta F  \leftarrow  ({x'}_i - x_i)*c_i$ \;
\For{$j \leftarrow 1$ to $n$}
{
  $\Delta$Sum $\leftarrow ({x'}_i - x_i)*q_{ij} $ \;
  newSum $\leftarrow Sum(x, j) + \Delta$Sum\;
  \eIf{$\Delta$Sum $> 0$ and newSum $> 0$}
  {
     $\Delta F  \leftarrow \Delta F$ + min($\Delta$Sum, newSum) \;
  }
  {
     \If{$\Delta Sum < 0$ and $Sum(x,j) > 0$}
     {
       $\Delta F  \leftarrow \Delta F - $min($-\Delta$Sum, $Sum(x,j)$) \;
     }
  }
}
\Return $\Delta F$ \;
\end{algorithm}

\subsection{Flip-float move based coordinate method}
Based on the {\it Flip-float} neighborhood, we get a coordinate method as described in Algorithm \ref{alg:coordinated} which alternatively uses {\it Flip-x-Float-y}  and  {\it Flip-y-Float-x} moves to improve a solution. Starting from a given initial solution, the algorithm first chooses the optimal $y$ for the given $x$. Then it progressively uses the {\it Flip-x-float-y} moves to improve the incumbent solution.  At each search step, the algorithm scans all possible {\it Flip-x-Float-y} moves. Once it encounters an improving move with $\Delta${\it Flip-x-Float-y} $(i) > 0$, it performs that move. If all {\it Flip-x-Float-y} moves can not improve the incumbent solution, the algorithm tries to improve the incumbent solution using the {\it Flip-y-Float-x} move in a similar manner as described above. If the solution has been improved by {\it Flip-y-Float-x} move, it will go back to the beginning and tries to improve it using the {\it Flip-x-Float-y} moves again. The algorithm terminates when the solution can not be improved by both {\it Flip-x-Float-y} and {\it Flip-y-Float-x} moves.
\begin{algorithm}
\caption{The flip-float move based coordinate method}
\label{alg:coordinated}
\KwIn{An initial solution $(x, y)$}
\KwOut{A locally optimal solution w.r.t. flip-float move}
\Repeat{the solution can not be improved by both Flip-x-Float-y and Flip-y-Float-x moves}{
Initialize $Sum(x, j)$ array according to Eq.\ref{eq:sumxj}\;
$y \leftarrow y^*(x)$ \;
\For{$i \leftarrow 1 $ to $m$}
{
   Determine $\Delta \textrm{\it Flip-x-Float-y}(i)$ using Algorithm \ref{alg:deltaF} \;
   \If{$\Delta$Flip-x-Float-y $(i) > 0$}
   {
     $x_i \leftarrow 1 - x_i$ \;
     Update $Sum(x,j)$ array \;
     $y \leftarrow y^*(x)$ \;
     goto Step 4 \;
   }
}
Initialize $Sum(y, i)$ array according to Eq.\ref{eq:sumyi}\;
$x \leftarrow x^*(y)$ \;
\For{$j \leftarrow 1 $ to $n$}
{
   Determine $\Delta \textrm{\it Flip-y-Float-x}(j)$ similar to Algorithm \ref{alg:deltaF} \;
   \If{$\Delta$Flip-y-Float-x $(j) > 0$}
   {
     $y_j \leftarrow 1 - y_j$ \;
     Update $Sum(y,i)$ array \;
     $x \leftarrow x^*(y)$ \;
     goto Step 15  \;
   }
}}
\Return $(x,y)$
\end{algorithm}

\subsection{Hybrid method}
The one-flip move based tabu search algorithm  and the flip-float move based coordinate method represent two  kinds of algorithmic paradigms. For the one-flip move based tabu search algorithm, each move is simple and fast, but the tabu search strategy is very powerful, making it possible to escape from small local minimum traps and  iterate for large number of steps . In the flip-float move based coordinate method, each move examines a large number of candidate solutions and therefore is more powerful and more expensive, but the coordinate ascent strategy is simple and allows only a few number of iterations in each local search.

The hybrid method integrates the one-flip tabu search algorithm and the flip-float coordinate method to yield the schema given in Algorithm \ref{alg:hybrid}. The integrating strategy is simple. It alternatively uses the one-flip tabu search algorithm and the flip-float coordinate method to improve the incumbent solution, until no improvement can be achieved using both methods.
\begin{algorithm}
\caption{The hybrid algorithm}
\label{alg:hybrid}
\KwIn{An initial solution $(x, y)$}
\KwOut{A best-found solution}
$\lambda \leftarrow  1$  \;
\While{$\lambda = 1$}
{
   $\lambda \leftarrow 0$ \;
   $(x,y) \leftarrow$ Improve the solution $(x,y)$ using the one-flip move based tabu search algorithm \;
   $(x,y) \leftarrow$ Improve the solution $(x,y)$ using the flip-float move based coordinate method \;
   \If{the solution was improved by the flip-float move based coordinate method}
   {
     $\lambda \leftarrow 1 $ \;
   }
}
\Return{$(x,y)$}
\end{algorithm}

\section{Experimental analysis}
This section provides experimental results of the described  algorithms on a set of 85 benchmark instances. We analyze their computational statistics in the aim of assessing their performance and disclosing their individual properties.

\subsection{Test instances and experimental protocol}
We adopt the standard testbed generated by Karapetyan and Punnen~\cite{kp} in our experiments. This testbed consists of 85 test instances which can be classified into
five categories according to their application background: Random instances, Max Clique instances, Max Induced Subgraph instances, Max Cut instances and Matrix Factorization instances. Each category contains three kinds of instances: small instances of size from $20 \times 50$ to $50 \times 50$, medium instances of size from $200 \times 1000$ to $1000 \times 1000$, and large instances of size from $1000\times 5000$ to $5000 \times 5000$. A detailed description of the problem generator can be found in \cite{kp}. All these instances are available in public from the website \url{http://www.sfu.ca/~dkarapet/}{}.

The algorithms are coded in C++ and complied using GNU GCC. All the computational experiments are carried out on a PC with two 3.1GHZ Intel Xeon E5-2687W CPUs and 128G Memory. The computer can run 32 computing threads at the same time, so we use multiple threads to compute multiple instances. No parallel computing techniques are used.

The algorithms are run in a multi-start fashion with randomly generated initial solutions to allow diversification. A time limit is set for each small, medium and large instance respectively at 100, 1000 and 10000 seconds. The experiments are carried out without special tuning of the parameters, i.e., all the parameters used in the algorithms are fixed for all instances considered. To capture the performance of each algorithm on each instance, we record the following values: the best-found solution, the number of initial solutions tried, the number of times the best-found solution is repeatedly hit, and the average elapsed time for detecting the best-known solution (calculated as dividing the number of hit times into the elpased time the best-found solution is last detected).

\subsection{Computational results and discussion}
Tables \ref{tab:small}, \ref{tab:median} and \ref{tab:large} respectively present the computational statistics of the three algorithms on the small, medium and large instances. In each table, columns 1 and 2 give the instance name and the previous best-known solution obtained by Karapetyan and Punnen~\cite{kp}.  Columns 3-14 report the computational statistics of the three algorithms: the deviation between our best-found solution and the previous best-known solution is listed in column 2 (Deviation), the number of initial solutions tried (\#Init), the number of times our best-found solution is repeatedly detected (\#Hit), and the average time needed to reach the best-found solution (Time).

Table \ref{tab:small} reports the computational statistics on the 35 small instances. For all the instances, each algorithm can reach the best-known solutions efficiently and consistently. For each small instance, the needed computing time is usually less than 1 millisecond. We also conjecture that for all these 35 small instances the current best-known solutions are already optimal. (Each of them usually is repeatedly hit more than 100,000 times.)

Table \ref{tab:small} also reveals some individual properties of each algorithm. For the 7 Biclique instances, the performance of the hybrid method and the flip-float coordinate ascent method is roughly the same. They have significantly higher success rate (\#Hit/\#Init) than the one-flip tabu search algorithm. For the other 28 instances, the performance of the hybrid method and the one-flip tabu search algorithm are at the same level and are better than the flip-float coordinate method in terms of success rate (approximately 100\% v.s. 50\%). These observations show that the hybrid method inherits good properties from both the one-flip tabu search algorithm and the flip-float coordinate method.

\begin{sidewaystable}[htbp]
  \centering
  \caption{Computational statistics of the three algorithms on the 35 small instances}
  \label{tab:small}%
  \scriptsize{
    \begin{tabular}{l|l|llll|llll|llll}
    \toprule
    \multirow{2}[0]{*}{Instance Name}& \multirow{2}[0]{*}{Best-known} & \multicolumn{4}{c|}{Hybrid Method} & \multicolumn{4}{c|}{One-flip Tabu Search} &\multicolumn{4}{c}{Flip-float coordinate method}\\
    \cline{3-14}
          &       & Deviation & \#Init & \#Hit & Time (ms)  & Deviation & \#Init & \#Hit & Time (ms)  & Deviation & \#Init & \#Hit & Time (ms) \\
    \hline
    Biclique20x50 & 18341 & 0     & 178594 & 178351 & 0     & 0     & 384859 & 1623  & 61    & 0     & 1116063 & 280506 & 0 \\
    Biclique25x50 & 24937 & 0     & 152478 & 145875 & 0     & 0     & 380436 & 1838  & 54    & 0     & 812083 & 782547 & 0 \\
    Biclique30x50 & 27887 & 0     & 150193 & 18399 & 5     & 0     & 348283 & 1464  & 68    & 0     & 685472 & 156750 & 0 \\
    Biclique35x50 & 32515 & 0     & 117354 & 103561 & 0     & 0     & 298058 & 4914  & 20    & 0     & 414726 & 118298 & 0 \\
    Biclique40x50 & 33027 & 0     & 94380 & 52725 & 1     & 0     & 237547 & 3041  & 32    & 0     & 420982 & 216525 & 0 \\
    Biclique45x50 & 37774 & 0     & 79341 & 78924 & 1     & 0     & 226604 & 1950  & 51    & 0     & 288844 & 288844 & 0 \\
    Biclique50x50 & 30124 & 0     & 84527 & 30712 & 3     & 0     & 213764 & 1240  & 80    & 0     & 337932 & 337431 & 0 \\
    BMaxCut20x50 & 9008  & 0     & 181658 & 105868 & 0     & 0     & 281593 & 121145 & 0     & 0     & 906202 & 322188 & 0 \\
    BMaxCut25x50 & 10180 & 0     & 176712 & 104961 & 0     & 0     & 251272 & 118318 & 0     & 0     & 647870 & 289626 & 0 \\
    BMaxCut30x50 & 13592 & 0     & 149703 & 114921 & 0     & 0     & 190134 & 145242 & 0     & 0     & 490369 & 119412 & 0 \\
    BMaxCut35x50 & 14024 & 0     & 138000 & 52490 & 1     & 0     & 172827 & 65710 & 1     & 0     & 377840 & 33659 & 2 \\
    BMaxCut40x50 & 17610 & 0     & 115954 & 48123 & 2     & 0     & 147998 & 57850 & 1     & 0     & 310749 & 15441 & 6 \\
    BMaxCut45x50 & 15252 & 0     & 106119 & 37982 & 2     & 0     & 127375 & 45431 & 2     & 0     & 249277 & 30217 & 3 \\
    BMaxCut50x50 & 19580 & 0     & 95127 & 33228 & 3     & 0     & 118656 & 40511 & 2     & 0     & 208773 & 32492 & 3 \\
    MatrixFactor20x50 & 114   & 0     & 221042 & 221042 & 0     & 0     & 286354 & 286354 & 0     & 0     & 670962 & 583384 & 0 \\
    MatrixFactor25x50 & 127   & 0     & 197375 & 197375 & 0     & 0     & 253512 & 253512 & 0     & 0     & 467549 & 438994 & 0 \\
    MatrixFactor30x50 & 148   & 0     & 158794 & 158794 & 0     & 0     & 204559 & 204559 & 0     & 0     & 366402 & 23789 & 4 \\
    MatrixFactor35x50 & 139   & 0     & 146580 & 146580 & 0     & 0     & 190794 & 190794 & 0     & 0     & 285900 & 166926 & 0 \\
    MatrixFactor40x50 & 210   & 0     & 135328 & 135328 & 0     & 0     & 175257 & 175257 & 0     & 0     & 220070 & 41211 & 2 \\
    MatrixFactor45x50 & 191   & 0     & 119706 & 119706 & 0     & 0     & 155124 & 155124 & 0     & 0     & 194567 & 74289 & 1 \\
    MatrixFactor50x50 & 217   & 0     & 108330 & 108330 & 0     & 0     & 144053 & 144053 & 0     & 0     & 158088 & 20158 & 4 \\
    MaxInduced20x50 & 6983  & 0     & 246453 & 208710 & 0     & 0     & 314245 & 263979 & 0     & 0     & 758467 & 485882 & 0 \\
    MaxInduced25x50 & 8275  & 0     & 186863 & 118483 & 0     & 0     & 244527 & 152919 & 0     & 0     & 540122 & 112608 & 0 \\
    MaxInduced30x50 & 10227 & 0     & 155943 & 150971 & 0     & 0     & 198937 & 192187 & 0     & 0     & 379252 & 106475 & 0 \\
    MaxInduced35x50 & 11897 & 0     & 156777 & 156777 & 0     & 0     & 200934 & 200934 & 0     & 0     & 343184 & 285156 & 0 \\
    MaxInduced40x50 & 14459 & 0     & 145671 & 145671 & 0     & 0     & 185726 & 185726 & 0     & 0     & 238934 & 229789 & 0 \\
    MaxInduced45x50 & 13247 & 0     & 108385 & 107353 & 0     & 0     & 138255 & 136921 & 0     & 0     & 200974 & 52996 & 1 \\
    MaxInduced50x50 & 15900 & 0     & 105508 & 105292 & 0     & 0     & 131545 & 131239 & 0     & 0     & 153775 & 102837 & 0 \\
    Rand20x50 & 13555 & 0     & 240731 & 240731 & 0     & 0     & 317500 & 317500 & 0     & 0     & 657434 & 382499 & 0 \\
    Rand25x50 & 13207 & 0     & 170476 & 107664 & 0     & 0     & 214032 & 135163 & 0     & 0     & 485003 & 111240 & 0 \\
    Rand30x50 & 15854 & 0     & 180159 & 180159 & 0     & 0     & 230848 & 230848 & 0     & 0     & 334042 & 269158 & 0 \\
    Rand35x50 & 14136 & 0     & 147259 & 145766 & 0     & 0     & 184884 & 183014 & 0     & 0     & 276747 & 64591 & 1 \\
    Rand40x50 & 18778 & 0     & 136871 & 136871 & 0     & 0     & 170684 & 170684 & 0     & 0     & 214272 & 116800 & 0 \\
    Rand45x50 & 22057 & 0     & 123368 & 123366 & 0     & 0     & 160725 & 160724 & 0     & 0     & 169020 & 25816 & 3 \\
    Rand50x50 & 23801 & 0     & 117483 & 117483 & 0     & 0     & 148970 & 148970 & 0     & 0     & 147063 & 123409 & 0 \\
    \bottomrule
    \end{tabular}%
    }
\end{sidewaystable}%

Table \ref{tab:median} gives the computational statistics of the three algorithm on the 25 medium sized instances.  The hybrid method and the one-flip tabu search algorithm successfully improve the previous best-known solutions on 11 instances and 10 instances respectively. The flip-float coordinate method fails to reach the previous best-known solutions on most instances within the given time limit. However, on the 5 Biclique instances where the one-flip tabu search performance is relatively poor, the flip-float coordinate method can improve the previous best-known solutions on 4 instances, demonstrating that it has some complementary property with respect to the one-flip tabu search algorithm. Compared with the one-flip tabu search method, the hybrid method usually find better solutions on the 10 Biclique and BMaxCut instances. For the other 15 instances, the best-found solutions are the same on each instance. However, the hybrid method is generally more robust because it usually has a smaller \#Init value but a larger \#Hit value.
\begin{sidewaystable}[htbp]
  \centering
  \caption{Computational statistics of the three algorithms on the 25 medium instances}
  \label{tab:median}%
  \scriptsize{
    \begin{tabular}{l|l|llll|llll|llll}
    \toprule
    \multirow{2}[0]{*}{Instance Name}& \multirow{2}[0]{*}{Best-known} & \multicolumn{4}{c|}{Hybrid Method} & \multicolumn{4}{c|}{One-flip Tabu Search} &\multicolumn{4}{c}{Flip-float coordinate method}\\
    \cline{3-14}
          &       & Deviation & \#Init & \#Hit & Time (s)  & Deviation & \#Init & \#Hit & Time (s)  & Deviation & \#Init & \#Hit & Time (s) \\
    \hline
    Biclique200x1000 & 2150201 & -3360 & 3653  & 3653  & 0     & -1943566 & 12904 & 8     & 86    & 0     & 6821  & 1     & 917 \\
    Biclique400x1000 & 4051884 & -171828 & 1342  & 1342  & 1     & -3748430 & 6676  & 1     & 129   & -5669 & 844   & 1     & 370 \\
    Biclique600x1000 & 5465191 & -44342 & 657   & 5     & 179   & -4782833 & 3603  & 1     & 788   & -11303 & 453   & 43    & 23 \\
    Biclique800x1000 & 6651165 & -23568 & 443   & 443   & 2     & -6066520 & 2352  & 1     & 66    & 36602 & 291   & 126   & 8 \\
    Biclique1000x1000 & 8601552 & 64393 & 329   & 158   & 6     & -3453555 & 2562  & 1     & 170   & 47798 & 255   & 255   & 4 \\
    BMaxCut200x1000 & 616810 & -2634 & 3521  & 1     & 144   & -10724 & 7637  & 1     & 695   & -2534 & 5818  & 1     & 745 \\
    BMaxCut400x1000 & 940944 & 8858  & 1729  & 1     & 879   & 9396  & 2334  & 1     & 515   & -2028 & 1182  & 1     & 790 \\
    BMaxCut600x1000 & 1520214 & -282972 & 1089  & 1     & 241   & -281540 & 1371  & 1     & 982   & -309568 & 456   & 1     & 473 \\
    BMaxCut800x1000 & 1215160 & 329390 & 756   & 1     & 667   & 327402 & 1020  & 1     & 574   & 288664 & 216   & 1     & 393 \\
    BMaxCut1000x1000 & 1771726 & 41174 & 599   & 1     & 818   & 40076 & 807   & 1     & 265   & -11084 & 134   & 1     & 650 \\
    MatrixFactor200x1000 & 6283  & 0     & 7340  & 130   & 8     & 0     & 9902  & 142   & 7     & -2    & 7524  & 1     & 947 \\
    MatrixFactor400x1000 & 9862  & 0     & 2862  & 899   & 1     & 0     & 3869  & 558   & 2     & -21   & 1546  & 1     & 146 \\
    MatrixFactor600x1000 & 12898 & 4     & 1950  & 355   & 3     & 4     & 2556  & 266   & 4     & -20   & 601   & 1     & 830 \\
    MatrixFactor800x1000 & 15437 & 29    & 1247  & 138   & 7     & 29    & 1602  & 95    & 10    & -42   & 289   & 1     & 4 \\
    MatrixFactor1000x1000 & 18792 & 21    & 1122  & 47    & 21    & 21    & 1510  & 25    & 40    & -9    & 158   & 1     & 270 \\
    MaxInduced200x1000 & 513081 & 0     & 5041  & 252   & 4     & 0     & 8605  & 241   & 4     & 0     & 4310  & 29    & 34 \\
    MaxInduced400x1000 & 777028 & 0     & 2401  & 91    & 11    & 0     & 3559  & 92    & 11    & -74   & 861   & 1     & 423 \\
    MaxInduced600x1000 & 973711 & 0     & 1265  & 456   & 2     & 0     & 1953  & 93    & 11    & -1676 & 354   & 1     & 450 \\
    MaxInduced800x1000 & 1204745 & 788   & 1034  & 9     & 107   & 788   & 1530  & 15    & 64    & -1009 & 169   & 1     & 846 \\
    MaxInduced1000x1000 & 1414743 & 879   & 828   & 77    & 13    & 879   & 1288  & 60    & 17    & -1376 & 110   & 1     & 118 \\
    Rand200x1000 & 612947 & 0     & 4908  & 84    & 12    & 0     & 8828  & 50    & 20    & 0     & 3511  & 18    & 55 \\
    Rand400x1000 & 951950 & 0     & 2256  & 565   & 2     & 0     & 3761  & 277   & 4     & 0     & 762   & 1     & 257 \\
    Rand600x1000 & 1345690 & 58    & 1555  & 84    & 12    & 58    & 2418  & 30    & 33    & -375  & 299   & 1     & 627 \\
    Rand800x1000 & 1604746 & 179   & 1192  & 51    & 19    & 179   & 1713  & 31    & 32    & -878  & 161   & 1     & 577 \\
    Rand1000x1000 & 1828902 & 1334  & 963   & 121   & 8     & 1334  & 1413  & 66    & 15    & -4348 & 92    & 1     & 23 \\
    \bottomrule
    \end{tabular}%
    }
\end{sidewaystable}%

\begin{sidewaystable}[htbp]
  \centering
  \caption{Computational statistics of the three algorithms on the 25 large instances}
  \label{tab:large}%
  \scriptsize{
    \begin{tabular}{l|l|llll|llll|llll}
    \toprule
    \multirow{2}[0]{*}{Instance Name}& \multirow{2}[0]{*}{Best-known} & \multicolumn{4}{c|}{Hybrid Method} & \multicolumn{4}{c|}{One-flip Tabu Search} &\multicolumn{4}{c}{Flip-float coordinate method}\\
    \cline{3-14}
          &       & Deviation & \#Init & \#Hit & Time (s)  & Deviation & \#Init & \#Hit & Time (s)  & Deviation & \#Init & \#Hit & Time (s) \\
    \hline
    Biclique1000x5000 & 38489329 & -130572 & 545   & 545   & 18    & -37445214 & 3690  & 1     & 6931  & -227214 & 192   & 3     & 1258 \\
    Biclique2000x5000 & 64124897 & 409442 & 166   & 111   & 90    & -62615159 & 1799  & 1     & 7941  & 551466 & 77    & 5     & 935 \\
    Biclique3000x5000 & 96735826 & -285670 & 83    & 83    & 121   & -94271142 & 1209  & 2     & 2017  & 832798 & 48    & 48    & 209 \\
    Biclique4000x5000 & 125690937 & 2241269 & 49    & 42    & 239   & -123684458 & 783   & 1     & 833   & 2534335 & 31    & 31    & 331 \\
    Biclique5000x5000 & 161974406 & 1541091 & 33    & 18    & 562   & -156946737 & 858   & 1     & 9289  & 1541091 & 23    & 23    & 447 \\
    BMaxCut1000x5000 & 6531128 & -34322 & 585   & 1     & 6509  & -157860 & 1417  & 1     & 9788  & -47464 & 331   & 1     & 1988 \\
    BMaxCut2000x5000 & 10085616 & 128586 & 385   & 1     & 1724  & 101198 & 619   & 1     & 7888  & -107548 & 59    & 1     & 3174 \\
    BMaxCut3000x5000 & 13505722 & 396512 & 283   & 1     & 9315  & 406968 & 440   & 1     & 537   & -71262 & 21    & 1     & 2515 \\
    BMaxCut4000x5000 & 16358716 & 623990 & 192   & 1     & 9248  & 617426 & 325   & 1     & 4832  & 23588 & 11    & 1     & 2082 \\
    BMaxCut5000x5000 & 19348266 & 739490 & 154   & 1     & 8464  & 746954 & 279   & 1     & 7697  & -76772 & 7     & 1     & 5211 \\
    MatrixFactor1000x5000 & 71470 & 0     & 1745  & 1     & 2138  & -8    & 2480  & 1     & 790   & -560  & 672   & 2     & 1793 \\
    MatrixFactor2000x5000 & 107939 & 94    & 1011  & 1     & 1569  & 66    & 1495  & 1     & 4880  & -461  & 127   & 1     & 2823 \\
    MatrixFactor3000x5000 & 143886 & 333   & 667   & 1     & 7435  & 319   & 1097  & 1     & 2910  & -327  & 44    & 1     & 9231 \\
    MatrixFactor4000x5000 & 178967 & 481   & 497   & 1     & 3325  & 502   & 782   & 1     & 657   & -346  & 19    & 1     & 6077 \\
    MatrixFactor5000x5000 & 210390 & 629   & 386   & 1     & 6368  & 644   & 657   & 1     & 8202  & -293  & 10    & 1     & 9774 \\
    MaxInduced1000x5000 & 5463868 & 446   & 1274  & 2     & 4997  & -323  & 2533  & 1     & 9941  & -2588 & 228   & 1     & 3293 \\
    MaxInduced2000x5000 & 8256468 & 8736  & 743   & 1     & 2839  & 7890  & 1517  & 1     & 4505  & -5374 & 45    & 1     & 8704 \\
    MaxInduced3000x5000 & 11070646 & 18920 & 481   & 1     & 4083  & 18672 & 1060  & 1     & 4256  & -13790 & 18    & 1     & 2867 \\
    MaxInduced4000x5000 & 13447665 & 47209 & 330   & 1     & 3117  & 45511 & 773   & 1     & 1302  & -5592 & 9     & 1     & 5955 \\
    MaxInduced5000x5000 & 15975303 & 44639 & 257   & 1     & 1721  & 44659 & 659   & 1     & 7019  & 1085  & 6     & 1     & 10896 \\
    Rand1000x5000 & 7182386 & 471   & 1334  & 1     & 7018  & -356  & 2858  & 1     & 4076  & -4323 & 219   & 1     & 1426 \\
    Rand2000x5000 & 11087619 & 8499  & 778   & 1     & 4404  & 6480  & 1635  & 1     & 6321  & -16577 & 46    & 1     & 2942 \\
    Rand3000x5000 & 14403998 & 29890 & 509   & 1     & 9829  & 28443 & 1173  & 1     & 8026  & -3511 & 18    & 1     & 8361 \\
    Rand4000x5000 & 18034574 & 33346 & 399   & 1     & 1004  & 29090 & 935   & 1     & 3159  & -12670 & 9     & 1     & 7399 \\
    Rand5000x5000 & 20946066 & 46635 & 336   & 1     & 1897  & 42930 & 721   & 1     & 2946  & -36659 & 5     & 1     & 2067 \\
    \bottomrule
    \end{tabular}%
    }
\end{sidewaystable}%

As shown in Table \ref{tab:large}, the hybrid method and the one-flip tabu search algorithm can improve most of the previous best-known solutions within the given time limit on the 25 large instances. The flip-float coordinate method performs best on the 5 Biclique instances;  improving 4 previous best-known solutions. However, it fails to improve 19 out of the other 20 instances. Compared with the one-flip tabu search algorithm, the hybrid method usually has a smaller \#Init value on each instance but is able to find better solutions on 20 out of all 25 instances.  In addition, we conjecture that for the 25 large instances,  most of the current best-found solutions are not optimal, because  the \#Hit value is usually 1. If we set the time limit to larger values, we may find better solutions.

In summary, our experimental results reveal that the one-flip tabu search algorithm performs well on most of the instances except for the Biclique instances (as opposed to our hybrid method incorporating the TS approach, which does well on all problems). The flip-float coordinate method is generally inferior to one-flip tabu search algorithm but performs significantly better on the Biclique instances . The hybrid method which integrates the one-flip tabu search algorithm and the flip-float coordinate ascent method shows better performance than both of them in the following three aspects: (1) It has good performance  regardless of the problem instance; (2) it is more robust; (3) its solution quality is generally better.

Since BBQP can be formulated as a mixed integer program (MIP), one can use any MIP solver as a heuristic by restricting its running time. To compare the performance of our heuristics to this ready-made heuristic algorithm we used the general purpose MIP solver CPLEX~\cite{ibm}.  The results are summarized in Tables \ref{tab:cplexsmall} and \ref{tab:cplexmedium}. The tables clearly establish that our metaheuristic algorithms possess significant advantages over this ready-made general purpose heuristic in terms of  efficiency as well as solution quality.  In our experiments,  we set the time limit of CPLEX the same as the time limit given to  the three metaheuristic algorithms. We also tested CPLEX as a heuristic by doubling this time limit. It may be noted that based on the experiments reported by Karapetyan and Punnen~\cite{kp} the CPLEX solver failed to find optimal solutions on most instances of size larger than $40 \times 50$ in 5 hours.  In our experiments, we set the CPLEX parameter MIPEmphasis  to 1, so that CPLEX put more emphasis on finding good feasible solutions and less emphasis on proof of optimality.

For small instances, CPLEX obtained optimal solutions within the allowed time (with proof of optimality) for 14 out of 35 problems and obtained the best known solutions for 27 out of 35 instances. For this class of instances, all our heuristics obtained the best known solutions  for all problems (including guaranteed optimal ones) and hit the first such solution in almost negligible time. By doubling the allowed running time,  CPLEX  matched the best known solutions for two additional instances.

For medium size instances, CPLEX obtained only the trivial solutions $x=0, y=0$ in 18 out of 25 cases and by doubling the allowed running time, it obtained non-trivial solutions for two additional instances. The quality of all these solutions is significantly lower than the quality obtained by our heuristics. For large scale instances, CPLEX reached allowed memory limit very quickly and hence we discontinued experiments with large instances.

\begin{table}[htbp]
  \centering
  \scriptsize{
    \begin{threeparttable}
    \caption{Computational results of the CPLEX on the small instances}
    \label{tab:cplexsmall}%
    \begin{tabular}{l|lll|lll}
    \toprule
    \multirow{2}[0]{*}{Instance Name} & \multicolumn{3}{c|}{Time Limit = 100 s} & \multicolumn{3}{c}{Time Limit = 200 s} \\
    \cline{2-7}
          & Value & Time (ms) & Optimality & Value & Time (ms) & Optimality \\
    \hline
    Biclique20x50 & 18341 & 225   & optimal & 18341 & 250   & optimal \\
    Biclique25x50 & 24937 & 490   & optimal & 24937 & 520   & optimal \\
    Biclique30x50 & 27887 & 1310  & optimal & 27887 & 1170  & optimal \\
    Biclique35x50 & 32515 & 1205  & optimal & 32515 & 1146  & optimal \\
    Biclique40x50 & 33027 & 3631  & optimal & 33027 & 3391  & optimal \\
    Biclique45x50 & 37774 & 10136 & optimal & 37774 & 10086 & optimal \\
    Biclique50x50 & 30124 & 44155 & optimal & 30124 & 42505 & optimal \\
    BMaxCut20x50 & 9008  & 4385  & optimal & 9008  & 4243  & optimal \\
    BMaxCut25x50 & 10180 & 47590 & optimal & 10180 & 49015 & optimal \\
    BMaxCut30x50 & 13592 & 100026 & feasible & 13592 & 200009 & feasible \\
    BMaxCut35x50 & 13084 & 100021 & feasible & 14024 & 200007 & feasible \\
    BMaxCut40x50 & 16916 & 100026 & feasible & 17392 & 200039 & feasible \\
    BMaxCut45x50 & 14422 & 100036 & feasible & 14662 & 200025 & feasible \\
    BMaxCut50x50 & 18766 & 100026 & feasible & 18788 & 200041 & feasible \\
    MatrixFactor20x50 & 114   & 30405 & optimal & 114   & 28345 & optimal \\
    MatrixFactor25x50 & 127   & 100030 & feasible & 127   & 200023 & feasible \\
    MatrixFactor30x50 & 148   & 100027 & feasible & 148   & 200039 & feasible \\
    MatrixFactor35x50 & 139   & 100027 & feasible & 139   & 200039 & feasible \\
    MatrixFactor40x50 & 210   & 100012 & feasible & 210   & 200039 & feasible \\
    MatrixFactor45x50 & 188   & 100012 & feasible & 191   & 200023 & feasible \\
    MatrixFactor50x50 & 213   & 100027 & feasible & 212   & 200039 & feasible \\
    MaxInduced20x50 & 6983  & 967   & optimal & 6983  & 936   & optimal \\
    MaxInduced25x50 & 8275  & 4197  & optimal & 8275  & 4212  & optimal \\
    MaxInduced30x50 & 10227 & 65254 & optimal & 10227 & 62322 & optimal \\
    MaxInduced35x50 & 11897 & 57142 & optimal & 11897 & 56659 & optimal \\
    MaxInduced40x50 & 14459 & 100027 & feasible & 14459 & 99075 & optimal \\
    MaxInduced45x50 & 13229 & 100027 & feasible & 13247 & 200023 & feasible \\
    MaxInduced50x50 & 15900 & 100027 & feasible & 15890 & 200023 & feasible \\
    Rand20x50 & 13555 & 1701  & optimal & 13555 & 1731  & optimal \\
    Rand25x50 & 13207 & 88546 & optimal & 13207 & 87656 & optimal \\
    Rand30x50 & 15854 & 100011 & feasible & 15854 & 200023 & feasible \\
    Rand35x50 & 14039 & 100011 & feasible & 14039 & 200023 & feasible \\
    Rand40x50 & 18778 & 100027 & feasible & 18778 & 200039 & feasible \\
    Rand45x50 & 22057 & 100011 & feasible & 22044 & 200039 & feasible \\
    Rand50x50 & 23801 & 100043 & feasible & 23720 & 200039 & feasible \\
    \bottomrule
    \end{tabular}
    \begin{tablenotes}
     \item[*] The results are obtained by CPELX with the following parameter settings: Threads = 1, MIPEmphasis = 1, MIPDisplay = 0, TreLim = 1000.
    \end{tablenotes}
    \end{threeparttable}
    }%
\end{table}%

\begin{table}[htbp]
  \centering
  \scriptsize{
    \begin{threeparttable}
    \caption{Computational results of the CPLEX on the medium instances}
    \label{tab:cplexmedium}%
    \begin{tabular}{l|lll|lll}
    \toprule
    \multirow{2}[0]{*}{Instance Name} & \multicolumn{3}{c|}{Time Limit = 1000 s} & \multicolumn{3}{c}{Time Limit = 2000 s} \\
    \cline{2-7}
          & Value &Time (s) & Optimality & Value & Time (s) & Optimality \\
    \hline
    Biclique200x1000 & 0     & 1002  & feasible & 0     & 2002  & feasible \\
    Biclique400x1000 & 0     & 1003  & feasible & 0     & 2003  & feasible \\
    Biclique600x1000 & 0     & 1005  & feasible & 0     & 2005  & feasible \\
    Biclique800x1000 & 0     & 1006  & feasible & 0     & 2006  & feasible \\
    Biclique1000x1000 & 0     & 1009  & feasible & 0     & 2009  & feasible \\
    BMaxCut200x1000 & 0     & 1002  & feasible & 0     & 2002  & feasible \\
    BMaxCut400x1000 & 0     & 1003  & feasible & 0     & 2004  & feasible \\
    BMaxCut600x1000 & 0     & 1005  & feasible & 0     & 2005  & feasible \\
    BMaxCut800x1000 & 0     & 1006  & feasible & 0     & 2007  & feasible \\
    BMaxCut1000x1000 & 0     & 1008  & feasible & 0     & 2008  & feasible \\
    MatrixFactor200x1000 & 0     & 1001  & feasible & 792   & 2001  & feasible \\
    MatrixFactor400x1000 & 292   & 1004  & feasible & 292   & 2003  & feasible \\
    MatrixFactor600x1000 & 0     & 1005  & feasible & 0     & 2004  & feasible \\
    MatrixFactor800x1000 & 0     & 1006  & feasible & 0     & 2006  & feasible \\
    MatrixFactor1000x1000 & 0     & 1008  & feasible & 0     & 2008  & feasible \\
    MaxInduced200x1000 & 61130 & 1002  & feasible & 61130 & 2002  & feasible \\
    MaxInduced400x1000 & 25642 & 1003  & feasible & 25642 & 2004  & feasible \\
    MaxInduced600x1000 & 4963  & 1005  & feasible & 4963  & 2005  & feasible \\
    MaxInduced800x1000 & 8940  & 1007  & feasible & 8940  & 2006  & feasible \\
    MaxInduced1000x1000 & 31662 & 1007  & feasible & 31662 & 2009  & feasible \\
    Rand200x1000 & 0     & 1002  & feasible & 40315 & 2001  & feasible \\
    Rand400x1000 & 0     & 1003  & feasible & 0     & 2004  & feasible \\
    Rand600x1000 & 120682 & 1005  & feasible & 120682 & 2005  & feasible \\
    Rand800x1000 & 0     & 1006  & feasible & 0     & 2006  & feasible \\
    Rand1000x1000 & 0     & 1008  & feasible & 0     & 2008  & feasible \\
    \bottomrule
    \end{tabular}
    \begin{tablenotes}
    \item[*]  The results are obtained by CPELX with the following parameter settings: Threads = 1, MIPEmphasis = 1, MIPDisplay = 0, TreLim = 1000.
    \end{tablenotes}
    \end{threeparttable}
  }%
\end{table}%

\subsection{Landscape analysis of problem instances}
The computational experiments demonstrate that the one-flip tabu search algorithm's performance is significantly worse on the Biclique and BMaxCut instances  than  on the other instances. In order to obtain some insight into this phenomena, we employ a fitness distance analysis on some representative instances to show their different landscape properties.

This analysis is performed on 5 medium instances: Biclique1000x1000, BMaxCut1000x1000, MatrixFactor1000x1000, MaxInduced1000x1000, Rand1000x1000. For each instance, we run the one-flip tabu search 1000 times from randomly generated starting points and obtain 1000 local minima. We get a sample point from each solution by calculating the gap between its objective value and the corresponding best-known value, and also the Hamming distance from the best-known solution.  For each instance, we then plot the 1000 sampled points  in a figure to estimate the distribution of local optima.

Fig.\ref{fig:landscape} gives the fitness distance scatter plots for the five representative instances. Fig.\ref{fig:landscape} shows that, for the Biclique and BMaxCut instances,  all the sampled local minima are far away from the corresponding best-known solutions. This  phenomenon suggests that the best-known solutions may be located in a very narrow valley and therefore hard to detect.  Especially for the Biclique instances, it seems that all the  sampled points  fall within two distinct regions. For the other three instances, the best-known solution is usually surrounded by many local optima whose objective value deteriorate with the increase of distance from the optimum. This kind of property makes the best-known solution relatively easier to detect, because the one-flip tabu search algorithm is able to escape from small local optimum traps and this property enables the algorithm to reach the best-known solution through a sequence of local minima with ascending objective value.

\begin{figure}
\caption{Fitness distance scatter plots on five representative instances}
\label{fig:landscape}
\subfigure[Biclique 1000x1000]{
\begin{tikzpicture}[scale=0.8]
	\begin{axis}[
       xmin = 0,
       ymin = 0,
       xlabel=Distance to optimum,
       ylabel=Fitness difference]
	\addplot[only marks, mark=+, color = blue]%
	table [x=dis, y=fgap] {Biclique1000x1000.rep};
	\end{axis}
\end{tikzpicture}
}
\subfigure[BMaxCut 1000x1000]{
\begin{tikzpicture}[scale=0.8]
	\begin{axis}[
       xmin = 0,
       ymin = 0,
       xlabel=Distance to optimum,
       ylabel=Fitness difference]
	\addplot[only marks, mark=+, color = blue]%
	table [x=dis, y=fgap] {BMaxCut1000x1000.rep};
	\end{axis}
\end{tikzpicture}
}

\subfigure[MatrixFactor 1000x1000]{
\begin{tikzpicture}[scale=0.8]
	\begin{axis}[
       xmin = 0,
       ymin = 0,
       xlabel=Distance to optimum,
       ylabel=Fitness difference]
	\addplot[only marks, mark=+, color = blue]%
	table [x=dis, y=fgap] {MatrixFactor1000x1000.rep};
	\end{axis}
\end{tikzpicture}
}
\subfigure[MaxInduced 1000x1000]{
\begin{tikzpicture}[scale=0.8]
	\begin{axis}[
       xmin = 0,
       ymin = 0,
       xlabel=Distance to optimum,
       ylabel=Fitness difference]
	\addplot[only marks, mark=+, color = blue]%
	table [x=dis, y=fgap] {MaxInduced1000x1000.rep};
	\end{axis}
\end{tikzpicture}
}

\subfigure[Rand 1000x1000]{
\begin{tikzpicture}[scale=0.8]
	\begin{axis}[
       xmin = 0,
       ymin = 0,
       xlabel=Distance to optimum,
       ylabel=Fitness difference]
	\addplot[only marks, mark=+, color = blue]%
	table [x=dis, y=fgap] {Rand1000x1000.rep};
	\end{axis}
\end{tikzpicture}
}
\end{figure}

\section{Conclusion}
In this work, we developed three heuristic algorithms for BBQP. The first algorithm employs a classic one-flip neighborhood and a simple tabu search strategy. The second one is based on a new powerful flip-float neighborhood  and a simple coordinate ascent method. The third algorithm integrates the first and second algorithms to create a hybrid method with the aim of inheriting good properties from each.

To assess the performance characteristics of the proposed algorithms, we have conducted systematic computational experiments on a set of 85 test instances. Our findings demonstrate that the hybrid method outperforms both the one-flip tabu search algorithm and the flip-float coordinate method: It generally finds better solutions than either component method on a wide range of test instances, and overall is more robust. In addition, the hybrid method is able to improve most of the previous best-known solutions on instances of medium and large size. We also compared our algorithms with CPLEX running in heuristic mode and all our algorithms generated superior outcomes in terms solution quality and running time compared to this ready-made heuristic approach. In addition to the development and comparison of heuristics, we also performed a landscape analysis to compare the relative difficulty levels of the benchmark instances. This study revealed interesting properties of the structure of these problems.

Our findings suggest the potential value of algorithmic enhancements for future research: (1) introducing 2-flip moves in the procedures currently studied; (2) combining the resulting methods to produce new hybrids; (3) using more advanced forms of tabu search for the direct 1-flip and 2-flip methods; (4) likewise using tabu search to exploit the flip-float neighborhood(in both 1-flip and 2-flip versions).

\end{document}